\documentclass[10pt,twocolumn,letterpaper]{article}

\usepackage[final]{cvpr}              

\usepackage{graphicx}
\usepackage{amsmath}
\usepackage{amssymb}
\usepackage{booktabs}
\usepackage{xcolor}
\definecolor{royblue} {HTML}{002484}
\usepackage{array}
\definecolor{bayrischblau}{RGB}{0, 141, 201}
\definecolor{cliptxt}{RGB}{225, 213, 231}
\definecolor{cliptxtborder}{RGB}{162, 131, 177}
\definecolor{clipimg}{RGB}{213, 232, 212}
\definecolor{clipimgborder}{RGB}{130, 179, 102}
\definecolor{clipmix}{RGB}{219, 223, 222}
\definecolor{clipmixborder}{RGB}{146, 155, 140}

\usepackage[pagebackref,breaklinks,colorlinks]{hyperref}

\usepackage[capitalize]{cleveref}
\crefname{section}{Sec.}{Secs.}
\Crefname{section}{Section}{Sections}
\Crefname{table}{Table}{Tables}
\crefname{table}{Tab.}{Tabs.}

\def\cvprPaperID{*****} 
\def\confName{CVPR}
\def\confYear{2022}

\begin{document}

\title{Less Is More: Linear Layers on CLIP Features as Powerful VizWiz Model}

\author{Fabian Deuser, Konrad Habel, Philipp J. Rösch,  Norbert Oswald\\
University of the Bundeswehr Munich\\
Institute for Distributed Intelligent Systems (VIS)\\
{\tt\small \{fabian.deuser,konrad.habel,philipp.roesch,norbert.oswald\}@unibw.de}
}
\maketitle

\begin{abstract}
Current architectures for multi-modality tasks such as visual question answering suffer from their high complexity. As a result, these architectures are difficult to train and require high computational resources. To address these problems we present a CLIP-based architecture that does not require any fine-tuning of the feature extractors. 
A simple linear classifier is used on the concatenated features of the image and text encoder. 
During training an auxiliary loss is added which operates on the answer types. The resulting classification is then used as an attention gate on the answer class selection. 
On the VizWiz 2022 Visual Question Answering Challenge we achieve \textbf{60.15 \%} accuracy on Task 1: Predict Answer to a Visual Question and AP score of \textbf{83.78 \%} on Task 2: Predict Answerability of a Visual Question.
\end{abstract}

\section{Introduction}
\label{sec:intro}
Many new architectures were developed in recent years and applied to data sets like VQAv2, GQA or VizWiz-VQA ~\cite{Gurari2018VizWizGC}. The VizWiz data set differs from other VQA data sets, because it has several issues in the data. Questions may not be answerable due to missing information in the images or the quality of the images may be extremely poor. Additionally the questions in the data set are not developed with a rigid set of rules, but are often colloquially. 
Last year's winning team used an extension of OSCAR. 
They added an optical character recognition (OCR) module to the model and introduced reference image matching.  
Their final system is an ensemble of 96 models. While ensembles are important to achieve competitive results, they are extremely costly to train.

Our approach focus on simplicity and usability. 
We use pre-trained image and text encoders from CLIP~\cite{clip} and train only a simple classification head.
CLIP is based on CNN~\cite{He2016DeepRL} respectively Vision Transformer~\cite{dosovitskiy2021an} for image encoding and a Transformer~\cite{vaswani2017attention} for text encoding. 
The CLIP model is pre-trained on 400 million image-text pairs with a contrastive objective to bring both modalities into the same embedding space. Since CLIP is trained with many samples, it also has OCR capabilities~\cite{clip}.

\begin{figure}[t]
\centering
  \includegraphics{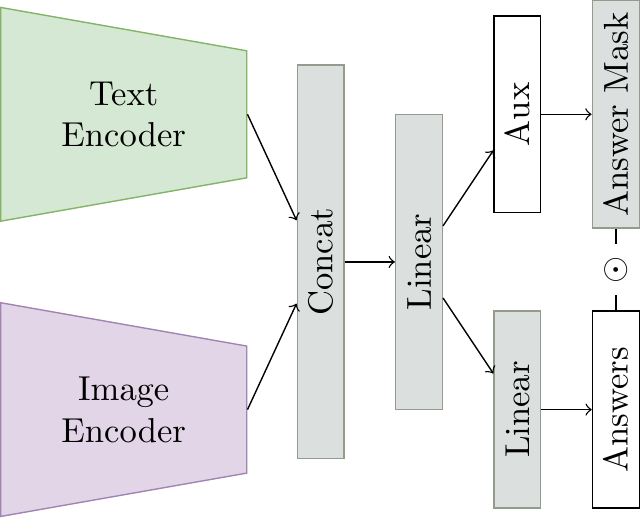}
\caption{Our architecture for the VizWiz Challenge 2022.}
\label{fig:clip}
\end{figure}

\begin{table*}[t]
\centering
\begin{tabular}{@{}lrrrr@{}}
\toprule
& \multicolumn{2}{c}{\textbf{VQA} [Acc.]}  & \multicolumn{2}{c}{\textbf{Answerability} [AP]} \\
Model           & test-dev      & test-std    & test-dev    & test-std  \\ \midrule 
{[R]} RN50x64         & $60.73$ \%           & $59.40$ \%                 & $82.74$ \%                    & $82.54$ \% \\ 
{[V]} ViT-L/14@336px  & $60.66$ \%           & $59.01$ \%                 & $83.50$ \%                    & $82.86$ \% \\ 
{[E]} Ensemble        & $61.64$ \%           & $60.15$ \%                 & $84.13$ \%                    & $83.78$ \% \\ 
\bottomrule 
\end{tabular}
\caption{Results in the VizWiz 2022 challenge.}
\label{tab:results}
\end{table*}

\section{Methodology}
\label{sec:methodology}

The contribution of this paper is divided into \textit{(i)} creating a suitable vocabulary for the classification task, \textit{(ii)} using CLIP features with linear layers for VQA, and \textit{(iii)} introducing an answer type gate to create a learnable masking. 

\paragraph{Answer Vocabulary.}
The selection of appropriate answers has a major impact on the accuracy that can be achieved. 
Therefore, in this approach, the most common answer that returns the highest score per image-question pair is greedily selected. If this selection yields in several answers, the answer which appears most often in the whole training set is used. 
In case of a tie, the pairwise Levenshtein distance is used to find the answer that is most representative to all others.
With this selection process the remaining number of answer candidates for training decreases to 5726.

\paragraph{CLIP-based Model.}
Previous CLIP-based models for VQA use the image encoder only~\cite{shen2022how} or generate prompts to match answers to questions~\cite{song2022clip}. Our approach utilises both image and text encoder. 
The resulting features are concatenated and passed to linear layers with layer normalisation and a high dropout value $(0.5)$. As shown in~\Cref{fig:clip} answer types as well as the answers are predicted using an additional linear layer. Image size of the visual encoder is 448x448 for RN50x64 and 336x336 for ViT-L/14@336px. In both cases the linear classifier is trained using cross entropy loss with rotation as image augmentation. We train only the additional linear classifier and use the pre-trained CLIP model as image and text encoder. The CLIP part remains frozen and is not trained on the VizWiz data set, which allows fast and efficient training without large computational resources.
\paragraph{Answer Type Gate.}
We also introduce an auxiliary loss for answer type prediction.
This loss helps to learn an answer masking for the eight answer types ``other", ``numbers", ``yes", ``no", ``color", ``unsuitable" and ``unanswerable". The answer types are retrieved by regular expression matching from the best selected answer per image-question pair.
The learned predictions for the answer types are linearly projected to a vector with the same dimension (5726) as the number of possible answer classes. 
After a sigmoid layer this vector is multiplied with the logits of the answer vocabulary. 
This enables to mask answers that do not correspond to the current answer type during inference. Both cross entropy losses, of the intermediate answer type prediction and the final answer classification, are weighted equally.

\section{Conclusion}
\label{sec:conclusion}
Our approach focuses on lightweight training by keeping the pre-trained CLIP backbone frozen, while still maintaining good accuracy.  
The OCR capabilities of CLIP, the large amount of pre-training data, and the multi-modality make CLIP an excellent feature extractor for this task. 
Unlike previous publications, the text Transformer is also used from CLIP. 
Although it was trained on alt-texts, it could be shown that meaningful representations of the questions are extracted without any fine-tuning.
On the VizWiz VQA task we reach \textbf{59.40}~\% with a single model and \textbf{60.15}~\% with an ensemble of the RN50x64 and ViT-L/14@336px. On the answerability task we achieve \textbf{82.86}~\%  with a single model and \textbf{83.78}~\% with an ensemble.

\section*{Acknowledgement}
The authors gratefully acknowledge the computing time granted by the Institute for Distributed Intelligent Systems and provided on the GPU cluster Monacum One at the University of the Bundeswehr Munich.

{\small
\bibliographystyle{ieee_fullname}
\bibliography{egbib}
}

\end{document}